\documentclass{article}

\usepackage{spconf,amsmath,graphicx}
\usepackage{amsfonts}
\usepackage{bm, bbm}
\usepackage{color}
\usepackage{booktabs}
\usepackage{array}
\usepackage[caption=false, font=footnotesize]{subfig}
\usepackage{dblfloatfix}
\usepackage{url}
\usepackage{textcomp}
\usepackage{multirow}
\usepackage{cite}
\usepackage{theorem}

\newtheorem{theorem}{Theorem}[section]
\newtheorem{definition}[theorem]{Definition}
\newtheorem{proposition}[theorem]{Proposition}

\newcommand{\RR}{\ensuremath{\mathbb R}}

\newcommand{\Tr}{\ensuremath{\operatorname{Tr}}}

\title{Multilayer Clustered Graph Learning}

\name{Mireille El Gheche  \qquad  \qquad \quad Pascal Frossard}

\address{mireille.elgheche@epfl.ch \qquad pascal.frossard@epfl.ch \\ [0.25em]
	Ecole Polytechnique F\'ed\'erale de Lausanne (EPFL), LTS4, Lausanne, Switzerland
}

\begin{document}

\maketitle

\begin{abstract}
Multilayer graphs are appealing mathematical tools for modeling multiple types of relationship in the data. In this paper, we aim at analyzing multilayer graphs by properly combining the information provided by individual layers, while preserving the specific structure that allows us to eventually identify communities or clusters that are crucial in the analysis of graph data. To do so, we learn a clustered representative graph by solving an optimization problem that involves a data fidelity term to the observed layers, and a regularization pushing for a sparse and community-aware graph. We use the contrastive loss as a data fidelity term, in order to properly aggregate the observed layers into a representative graph. The regularization is based on a measure of graph sparsification called "effective resistance", coupled with a penalization of the first few eigenvalues of the representative graph Laplacian matrix to favor the formation of communities. The proposed optimization problem is nonconvex but fully differentiable, and thus can be  solved via the projected gradient method. Experiments show that our method leads to a significant improvement w.r.t. state-of-the-art multilayer graph learning algorithms for solving clustering problems.
\end{abstract}


\section{Introduction}
\label{sec:intro}

A multilayer graph is a set of undirected graphs that shares the same set of nodes, and each layer describes a different type of relationships between the nodes. 
For example, a social network system can be represented by a multilayer graph, where the nodes are the users, and each layer describes a different kind of relationship (geometric distance, relational information, behavioral relationships based on user actions or interests, etc). Multilayer graphs are considered in many machine learning and data mining tasks, including multi-view learning, processing, and clustering or community detection. We focus here on the multilayer graph clustering problem, where the goal is to assign each graph node (shared across different layers) to a cluster by taking into account the information given by the different layers.


Prior work on multilayer graph clustering 
can be broadly categorized into layer-wise methods and global methods. On the one hand, layer-wise clustering methods infer a separate cluster of nodes in each layer, and produce multiple potential community associations for each node \cite{Tagarelli2017, Wang2019}. Different approaches consist of a hierarchical multi-view clustering algorithm \cite{Bickel2004,Chierchia_neurips2019} or a classification algorithm defined on each view through a multi-view regularization \cite{Sindhwani2005}. These methods can be useful when a unified representation for the multiple views is not easy to find in the data. 

On the other hand, global methods process the entire multilayer graph as a whole, and obtain a consensus community structure. For example, one can combine information from multilayer graphs into a representative graph that is embedded into a low-dimensional space, where the goal is to assign each graph node to a cluster. The most straightforward way is to combine linearly different layers \cite{Tang_20012, Chen_Hero_2017}. However, these methods may not be able to capture the structure information present in each layer. In order to properly take into account the topology shared across layers, one can see the graph layers as points of a Grassman manifold \cite{Wang_TIP_2013} or as points in a symmetric positive definite manifold \cite{TSIPN_elgheche2019}. The latter algorithm has been proven to have a higher accuracy compared to \cite{Wang_TIP_2013} but also to have higher complexity.

In this paper, we address the above issues by learning a clustered 
representative graph while integrating the information of multiple views to offer an effective way to cluster nodes of a multi-layer graph. In particular, we propose a new approach that aims at learning a representative graph by combining three ideas, namely graph learning, graph sparsification, and node clustering. We formulate these tasks as an optimization problem involving a contrastive loss as a data fidelity term w.r.t. the observed layers, and two problem-specific regularization terms. The first term is a measure of effective resistance to favor the estimation of a sparse graph. The second term is a clustering-like regularization that aims at capturing the structure information scattered among the observed layers. 
The resulting optimization problem is efficiently solved via projected gradient descent. Experimental results show that the proposed approach achieves a better clustering performance compared to baseline multilayer graph clustering approaches, due to the effective combination of the graph layers leading to a good structured representative graph.


\section{Problem formulation}
\label{sec:problem_formulation}

Let $\mathcal{G}$ be a multilayer graph with a set $V$ of $N$ vertices shared across $S \geq 1$ layers of edges, defined as
\begin{equation}
\mathcal{G} = \big\{\mathcal{G}^s(V,E^s)\big\}_{1\le s\le S}.
\end{equation} 
For each layer $s\in\{1,\dots,S\}$, 
we denote by $W^s = [w^s_{i,j}] \in \{0,1\}^{N\times N}$ the adjacency matrix of $\mathcal{G}^s$. The degree of a vertex $i$ in the graph $\mathcal{G}^s$, denoted as $d^s(i)$, is defined as the sum of all the edges incident to $i$ in the graph $\mathcal{G}^s$. The degree matrix $D^s$ is then defined as
\begin{equation}
D^s_{i,j} = \begin{cases} d^s(i) \quad & \textup{if $i=j$} \\
0 \quad & \textup{otherwise.}
\end{cases}
\end{equation}
The Laplacian matrix of $\mathcal{G}^s$ is thus defined as 
\begin{equation}
L^s=D^s-W^s.
\end{equation}

In this paper, we address the problem of analyzing multilayer graphs and propose a method for learning a representative graph with a Laplacian matrix $L$ that combines the information provided by the different layers, while preserving the specific structure that allows us to identify communities or clusters in the graph. A possible way to build such a representative graph consists of learning a valid Laplacian matrix through the resolution of the following optimization problem
\begin{equation}
\label{eq:laplacian_learning}
\operatorname*{minimize}_{L\in \mathcal{C}} \sum_{s=1}^S \mathcal{J}(L;\mathcal{G}^s) + \mathcal{R}(L),
\end{equation}
where $\mathcal{J}(\cdot\,; \mathcal{G}_s)$ is a data-fidelity term with respect to layer $s$, $\mathcal{R}$ is a suitable regularization term, and $\mathcal{C}$ is the space of valid combinatorial graph Laplacian matrices defined as
\begin{equation}
\mathcal{C} = \{L\in\mathbb{R}^{N\times N} \,|\, \; L_{ij}=L_{ji}\leq 0, L_{ii}=-\sum_{j\neq i} L_{ij}\}.
\end{equation}


The main difficulty in solving Problem \eqref{eq:laplacian_learning} arises from the space of valid Laplacian matrices, which is difficult to handle by first-order optimization methods such as projected gradient descent. 
We thus reformulate the optimization problem presented in \eqref{eq:laplacian_learning} by replacing the space of Laplacian matrices $\mathcal{C}$ with the space of valid upper-triangular adjacency matrices, which boils down to the set of the nonnegative vectors $\RR_{+}^{N(N-1)/2}$. Since any $L \in \mathcal{C}$ is a symmetric matrix with a degree of freedom equal to $N(N-1)/2$, we can define a linear operator $\mathcal{L}$ that maps a nonnegative vector $w\in \RR_{+}^{N(N-1)/2}$ into a matrix $\mathcal{L}w\in\mathcal{C}$.
\begin{definition}{$\mathcal{L}\colon \RR^{(N-1)N/2}\rightarrow \RR^{N\times N}$ is the linear operator that converts the upper-triangular part of an adjacency matrix into the corresponding Laplacian matrix.
}
\end{definition}
Thanks to the operator $\mathcal{L}$, we can rewrite Problem \eqref{eq:laplacian_learning} as 
\begin{equation}
\label{eq:problem_reg}
\operatorname*{minimize}_{w\in\RR_{+}^{(N-1)N/2}}\; \sum_{s=1}^S \mathcal{J}(\mathcal{L}w;\mathcal{G}^s) + \underbrace{\gamma_{1} \mathcal{R}_{\rm eff}(\mathcal{L}w) + \gamma_{2} \mathcal{R}_{\rm com}(\mathcal{L}w)}_{\mathcal{R}(\mathcal{L}w)},
\end{equation}
where $\gamma_{1}>0$ and $\gamma_{2}>0$ control the regularizations strength. To model the data-fidelity term $\mathcal{J}$, we employ the contrastive loss detailed in Section \ref{sub:contrastive_loss}. For the regularization $\mathcal{R}$, we actually use two pieces of information: the term $\mathcal{R}_{\rm eff}$ discussed in Section \ref{sub:sparse} that penalises the effective resistance of the learned graph, and the term $\mathcal{R}_{\rm com}$ presented in Section \ref{sub:clustering} that favors the formation of communities in the learned graph.

\subsection{Contrastive loss}
\label{sub:contrastive_loss}
In order to properly aggregate the observed layers into a representative graph, we employ the contrastive loss. It intends to maximize the likelihood of preserving neighborhoods of nodes present in each graph layer.
The contrastive loss was originally introduced in \cite{LeCunn2016_dimreduction,node2vec2016}, and used later for self-supervised learning \cite{khosla2020supervised, Chen2020ASF, He_2020_CVPR, Sohn_2016_npairloss, wu2018unsupervised} and for metric learning \cite{Weinberger2009}. 

Let $\mathcal{N}^s(i)$ be the set of nodes that appear in the neighboring of a node $i$. We define the constrastive loss for graph learning as
\begin{equation}
\label{eq:contrastive_emb}
\mathcal{J}(L; \mathcal{G}^s ) = \sum_{i=1}^N \sum_{j \in \mathcal{N}^s(i)} -\log\left( \dfrac{\exp\big( -L_{ij}\big)}{\sum_{k\neq i}\exp\big(-L_{ik}\big)} \right),
\end{equation}
where $w_{ij} = -L_{ij}$ is the weight on the edge $(i,j)$. The contrastive loss aims at learning the graph edge weights in such a way that the neighbors are pulled together and non-neighbors are pushed apart. The representative graph is thus estimated from the given layers $\{\mathcal{G}^s\}_{1\leq s \leq S}$ by using the neighboring connectivity information $\{\mathcal{N}^s\}_{1\leq s \leq S}$ only.

\subsection{Effective resistance}
\label{sub:sparse}

For improved interpretability and precise identification of the data structure, it is desirable to learn a graph with a significant degree of sparsity \cite{friedman_sparse_2008,Tarzanagh2017}. Hence, our first regularization is based on a measure of graph sparsification called "effective resistance".  The effective resistance is an interesting graph measure derived from the field of electric circuit analysis, where it is defined as the accumulated effective resistance between all pairs of vertices. In this paper, we use the effective resistance to measure the sparsification of a graph. The idea behind sparsification is to approximate (in some appropriate metric) a large graph by a sparse graph on the same set of vertices. Then, the sparse graph can be used as a proxy for the original graph with the advantage of cheaper computation and storage.

Given a graph Laplacian matrix $L\in\RR^{N\times N}$, the effective resistance $R_{ij}$ between a pair of nodes $i$ and $j$ is defined as
\begin{equation}
R_{ij} = v_i - v_j,
\end{equation}
where the vector $v=[\dots, v_i, \dots, v_j, \dots]^\top\in\RR^{N}$ is the solution to the equation
\begin{equation}\label{eq:vsol}
Lv = \delta_i - \delta_j,
\end{equation}
with $\delta_n$ being the $n$-th column of the $N\times N$ identity matrix. Note that all solutions to Eq.\ \eqref{eq:vsol} give the same value of $v_i - v_j$. Now, let us define the pseudoinverse of $L$ as
\begin{equation}
L^\dagger = (L + \mathbbm{1}\mathbbm{1}^\top/N)^{-1} - \mathbbm{1}\mathbbm{1}^\top/N,
\end{equation}
where $\mathbbm{1}=[1,\dots,1]^\top \in \RR^N$. Then, the effective resistance can be rewritten as \cite{klein1993}
\begin{equation}
R_{ij} = (\delta_i - \delta_j)^\top L^\dagger (\delta_i - \delta_j),
\end{equation}
and the total effective resistance amounts to \cite{ghosh2008}
\begin{equation}
R_{\rm tot} = \sum_{i < j} R_{ij} = N \Tr(L^\dagger) = N \sum_{n=2}^N \frac{1}{\lambda_n},
\end{equation}
where $0=\lambda_1<\lambda_2\le\dots\le\lambda_N$ are the eigenvalues of $L$. In this paper, we adopt a regularization that is proportional to the total effective resistance, namely
\begin{equation}
\mathcal{R}_{\rm eff}(L) = \sum_{n=K+1}^N \frac{1}{\lambda_n(L)},
\end{equation}
where $\lambda_n(L)$ denotes the $n$-th eigenvalue of $L$, and $K\ge1$ is equal to the number of communities chosen in the regularization discussed in Section \ref{sub:clustering}.

\subsection{Community regularization}
\label{sub:clustering}
While graph sparsification is widely used, especially in high-dimensional settings, it is not enough to learn a graph with a specific structure. Hence, our  
second regularization favors the formation of communities. The information structure about a graph can be concisely captured through the estimation of several properties. One of the most useful ways to do this has been by studying the eigenvalues of the Laplacian matrix.  By looking at the eigenvalues it is possible to get information about a graph that might otherwise be difficult to obtain. Spectral graph theory is one of the main tools to study the relationship between the eigenvalues and the structure of a graph. We will now recall two basic facts of spectral graph theory. Let $L$ be the Laplacian matrix of an undirected graph, along with the corresponding eigenvectors  $U=[u_1, \cdots, u_N]$ and eigenvalues $\Lambda= {\rm Diag}(\lambda_1, \cdots, \lambda_{N})$. 

\begin{definition}
A  connected  component  of  an  undirected  graph  is  a  connected subgraph such that there are no edges between vertices of the subgraph and vertices of the rest of the graph.
\end{definition}

\begin{proposition}
A graph has $K$ connected components if its vertex set can be partitioned
into $K$ disjoint subsets such that any two nodes belonging to different subsets are not connected. The eigenvalues of its Laplacian matrix are as follows
\begin{equation}
0 = \lambda_1 = \dots = \lambda_K \qquad 0 < \lambda_{K+1} \le \dots \le \lambda_N.
\end{equation}
\end{proposition}

To favor the learning of a graph with $K$ communities that are not necessarily disconnected, we propose to minimize the first $K$ eigenvalues of the corresponding Laplacian matrix, yielding the following regularization
\begin{equation}
\mathcal{R}_{\rm com}(L) = \sum_{n=1}^K \lambda_n^2(L),
\end{equation}
where $K\ge 1$, and $\lambda_n(L)$ denotes the $n$-th eigenvalue of $L$.

\section{Numerical results}
\label{sec:results}

In this section, we illustrate the advantages of incorporating structural information directly into learning a representative graph from multiple graph layers by solving Problem \eqref{eq:problem_reg}, which is fully differentiable and thus can be solved via projected gradient descent. We will carry on experiments with synthetic and real datasets having a multilayer graph representation.

\paragraph*{Datasets}

\begin{figure}[!h]
	\subfloat[Multilayer graph with $S=3$ layers. From the left to the right: layer 1, layer 2, and layer 3. The colors indicate the groundtruth clusters. \label{fig:gaussian_layers}]{\includegraphics[width=0.99\linewidth]{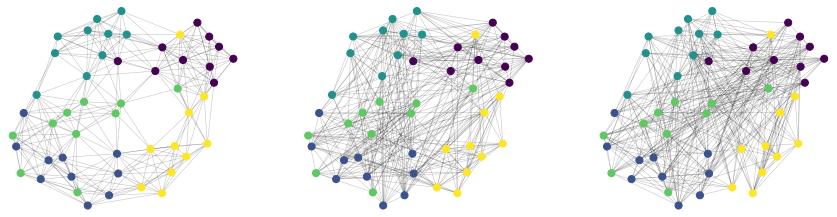}}
	\hfill
	\subfloat[Representative graphs. From the Left to the right: Arithmetic mean, Geometric mean \cite{TSIPN_elgheche2019}, Projection mean \cite{Dong_TSP_2014}, and the proposed method. Node coloring is the result of spectral clustering with $K=5$ on each estimated representative graph. \label{fig:gaussian_clustering}]{\includegraphics[width=0.99\linewidth]{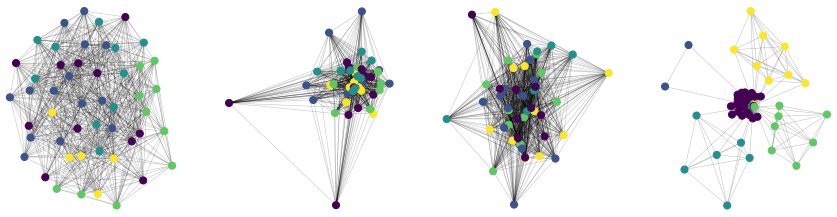}}
	\hfill
	\subfloat[Representative graphs. From the Left to the right: Arithmetic mean, Geometric mean \cite{TSIPN_elgheche2019}, Projection mean \cite{Dong_TSP_2014}, and the proposed method. Node coloring is the ground truth labels. Numerical evaluataions are reported in Table \ref{tab:synt}. \label{fid:gaussian_true}]{\includegraphics[width=0.99\linewidth]{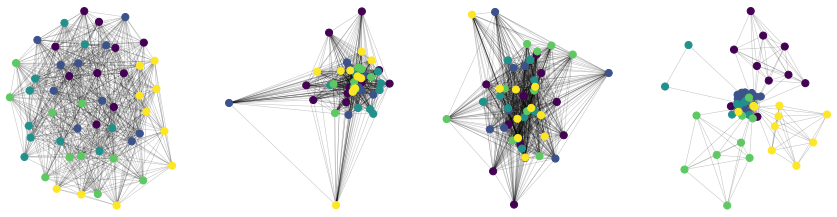}}
	\hfill
	\subfloat[Plots of the adjacency matrices. From the Left to the right: Arithmetic mean, Geometric mean \cite{TSIPN_elgheche2019}, Projection mean \cite{Dong_TSP_2014}, and the proposed method. Numerical evaluataions are reported in Table \ref{tab:synt}. \label{fig:adjacencies}]{\includegraphics[width=0.99\linewidth]{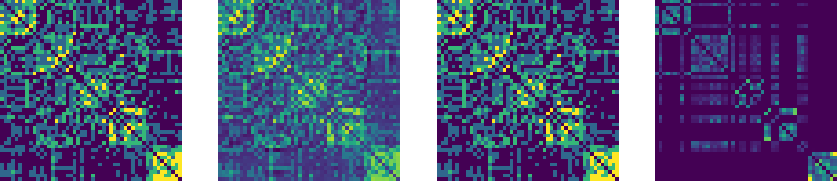}}
	\caption{Illustrative example of multilayer representative graph learning. The original multilayer graph is composed of three layers, which only provide a partial information on the clustering structure.}
	\label{fig:gaussian}
\end{figure}

Our first dataset is synthetic. It consists of $S=3$ point clouds of size $N=50$, each generated from a $2$-dimensional Gaussian mixture model with $K=5$ components having different means and covariance matrices. We build a 20-nearest-neighbor ($k$-NN) graph on each point cloud. The layer alignment is purely based on the clusters, as shown in Figure \ref{fig:gaussian_layers}. The goal with this dataset is to recover the five clusters (indicated by the node colors) of the 50 vertices using the three graph layers constructed from the three point clouds.

The second dataset consists of data collected at an anonymous university research department \cite{Rossi2015}. This dataset contains five layers that represent relationships between 61 employees (professors, postdoctoral, researchers, PhD students and administration staff). In this multilayer graph, considering the users as vertices in the graph, the authors of \cite{Rossi2015} construct five graphs by taking into account five different aspects: coworking, having lunch together, facebook relionship, offline friendship (having fun together), and co-authorship. The goal of this dataset is to cluster the researcher into groups (research groups), whose groundtruth is provided as an attribute per node.

\paragraph*{Comparison}

We use four criteria to measure the clustering performance: Accuracy, Purity, Normalized Mutual Information (NMI), and Rand Index (RI). We learn the representative graph of the multilayer graph using four different methods: the arithmetic mean (average), the projection mean in Grassman manifold \cite{Dong_TSP_2014}, the geometric mean in SPD manifold \cite{TSIPN_elgheche2019} and the proposed approach. Then, we perform spectral clustering \cite{Fortunato2010} on each learned graph, leading to the four indicators reported in Table \ref{tab:synt} and Table \ref{tab:social}. Among the results obtained with spectral clustering, the proposed approach achieves the best performance (in terms of Accuracy, Purity, NMI and RI). This empirically confirms that taking into account the structure of the graph carries a richer information than just aggregating the layers.  

Figure \ref{fig:gaussian_layers} gives an example of a multilayer graph, where the node colors are the groundtruth labels. We present the representative graph computed with four different approaches: arithmetic mean, geometric mean \cite{TSIPN_elgheche2019}, projection mean \cite{Dong_TSP_2014} and the proposed approach. Then we perform the spectral
clustering of the resulting representative (single-layer) graph, as detailed in Figure \ref{fig:gaussian_clustering}. The groundtruth labels are given in Figure \ref{fid:gaussian_true} for evaluation. As we can notice, the proposed approach is able to provide a structured and sparse graph with rich information between the clusters. This is validated with the adjacency matrices in Figure \ref{fig:adjacencies}, where the orderings of the vertices are made consistent with the groundtruth clusters. These plots provide a global view of a matrix where every non-zero entry in the matrix is represented by a colorful dot. As shown in Table \ref{tab:synt}, the clusters in the representative graph computed with the proposed approach are better than the clusters in the arithmetic mean, geometric mean \cite{TSIPN_elgheche2019}, and projection mean \cite{Dong_TSP_2014}. The reason for this is that, in these methods, there are no priors to model the structural information provided by the observation layers.

\begin{table}[t]
\centering
\caption{Synthetic data ($N=50$, $S=3$, $K=5$). \label{tab:synt}}
\begin{tabular}{ l   c  cc   c   c}
\toprule
Method & Accuracy & Purity & NMI & RI \\
\midrule	 
Arithmetic mean  &  0.58 & 0.60 & 0.48 & 0.31 \\
Geometric mean \cite{TSIPN_elgheche2019} &  0.60 & 0.64 & 0.52 & 0.35 \\
Projection mean  \cite{Dong_TSP_2014} & 0.58 & 0.6 & 0.48 & 0.31 \\
Proposed & 0.72 & 0.72 & 0.59 & 0.37 \\
\bottomrule	
\end{tabular}
\end{table}

\begin{table}[t]
\centering
\caption{Social data ($N=61$, $S=5$, $K=8$). \label{tab:social}}
\begin{tabular}{ l   c  cc   c   c}
\toprule
Method & Accuracy & Purity & NMI & RI \\
\midrule	 
Arithmetic mean  &  0.36 & 0.49 & 0.38 & 0.08 \\
Geometric mean \cite{TSIPN_elgheche2019} & 0.46 & 0.54 & 0.45 & 0.17 \\
Projection mean  \cite{Dong_TSP_2014} & 0.38 & 0.47 & 0.41 & 0.10 \\
Proposed & 0.50 & 0.56 & 0.38 & 0.14 \\
\bottomrule	
\end{tabular}
\end{table}

\section{Conclusion}
\label{sec:conclusion}

In this paper, we proposed a new optimization approach in order to estimate a community-based representative graph of a given multilayer graph. To do so, we have combined three ideas: graph learning, graph sparsification and node clustering. The resulting optimization model is well adapted to graph learning with sparse and community structure, as shown in Figure \ref{fig:gaussian}. This confirms that the proposed approach is capable to estimate a structural graph with rich information compared to competitor approaches.  

There are a number of extensions that we are currently investigating, such as the possibility to include a new modeling of
node features through a general approach to simultaneously aggregate the multilayer information with the graph data. Moreover, we are planning to extend our approach to compute the embedding matrices of each layer simultaneously with the representative graph.

\bibliographystyle{IEEEbib}
\bibliography{strings,biblio,biblio2}

\end{document}